\documentclass[a4paper,conference]{IEEEtran}
\usepackage{booktabs} 

\ifCLASSINFOpdf
  \usepackage[pdftex]{graphicx}
\else
\fi

\usepackage{color}
\usepackage{multirow}
\usepackage{amssymb}
\usepackage{amsmath}
\usepackage{bm}
\usepackage{upgreek}
\usepackage[numbers,sort&compress]{natbib}

\ifCLASSOPTIONcompsoc
  \usepackage[caption=false,font=normalsize,labelfont=sf,textfont=sf]{subfig}
\else
  \usepackage[caption=false,font=footnotesize]{subfig}
\fi
\begin{document}
%
\title{InsideBias: Measuring Bias in Deep Networks\\ and Application to Face Gender Biometrics}

 \author{\IEEEauthorblockN{Ignacio Serna, Alejandro Pe\~na, Aythami Morales, Julian Fierrez}
 \IEEEauthorblockA{School of Engineering, Universidad Autonoma de Madrid, Spain\\
 \{ignacio.serna, alejandro.penna, aythami.morales, julian.fierrez\}@uam.es}
 }


%


\maketitle

\begin{abstract}
This work explores the biases in learning processes 
based on deep neural network architectures. We analyze how
bias affects deep learning processes through a toy example using the
MNIST database and a case study 
in gender detection from face images. We employ two gender 
detection models based on popular deep neural networks. We
present a comprehensive analysis of bias effects when using
an unbalanced training dataset on the features learned by the
models. We show how bias impacts in the activations 
of gender detection models based on face images. We finally
propose InsideBias, a novel method to detect biased models.
InsideBias is based on how the models represent the information 
instead of how they perform, which is the normal practice in 
other existing methods for bias detection. Our strategy with 
InsideBias allows to detect biased models with very few samples
(only 15 images in our case study). Our experiments include 72K 
face images from 24K identities and 3 ethnic groups. 
\end{abstract}


%
\IEEEpeerreviewmaketitle

\section{Introduction}

Artificial Intelligence (AI) algorithms have an increasingly growing role in our
daily lives. These algorithms influence now many decision-making processes affecting
people’s lives in many important fields, e.g. social networks, forensics, health, and banking.
For example, some companies already use AI to predict credit risk, 
and some US states run prisoner details through AI systems
to predict the likelihood of recidivism when considering parole \cite{stone2016artificial}.

Face recognition algorithms are good examples of recent 
advances in AI. During the last ten
years, the accuracy of face recognition systems has increased 
up to 1000x (it is probably the biometric technology with the 
greatest investment nowadays). These face recognition algorithms 
are dominated by Deep Neural Network architectures, which 
are trained with huge amounts of data with 
little control over what is happening during training 
(focused on performance maximization). As a result, we 
have algorithms with excellent performance but quite opaque.

This trend in AI (excellent performance + low transparency) 
can be observed not only in face biometrics, but also in many 
other AI applications as well \cite{szegedy2013intriguing}. At this point, 
and despite the extraordinary advances in recognition performance, factors 
such as the lack of transparency, discrimination, and privacy issues 
are limiting many AI practical applications. As an example of these 
increasing concerns, in May 2019, the Board of Supervisors of San
Francisco banned the use of facial recognition software by the 
police and other agencies, and many others are considering or 
already have enacted legislation \cite{cook2019demographic}.

\begin{figure}[!t]
\centering
\includegraphics[width=85mm]{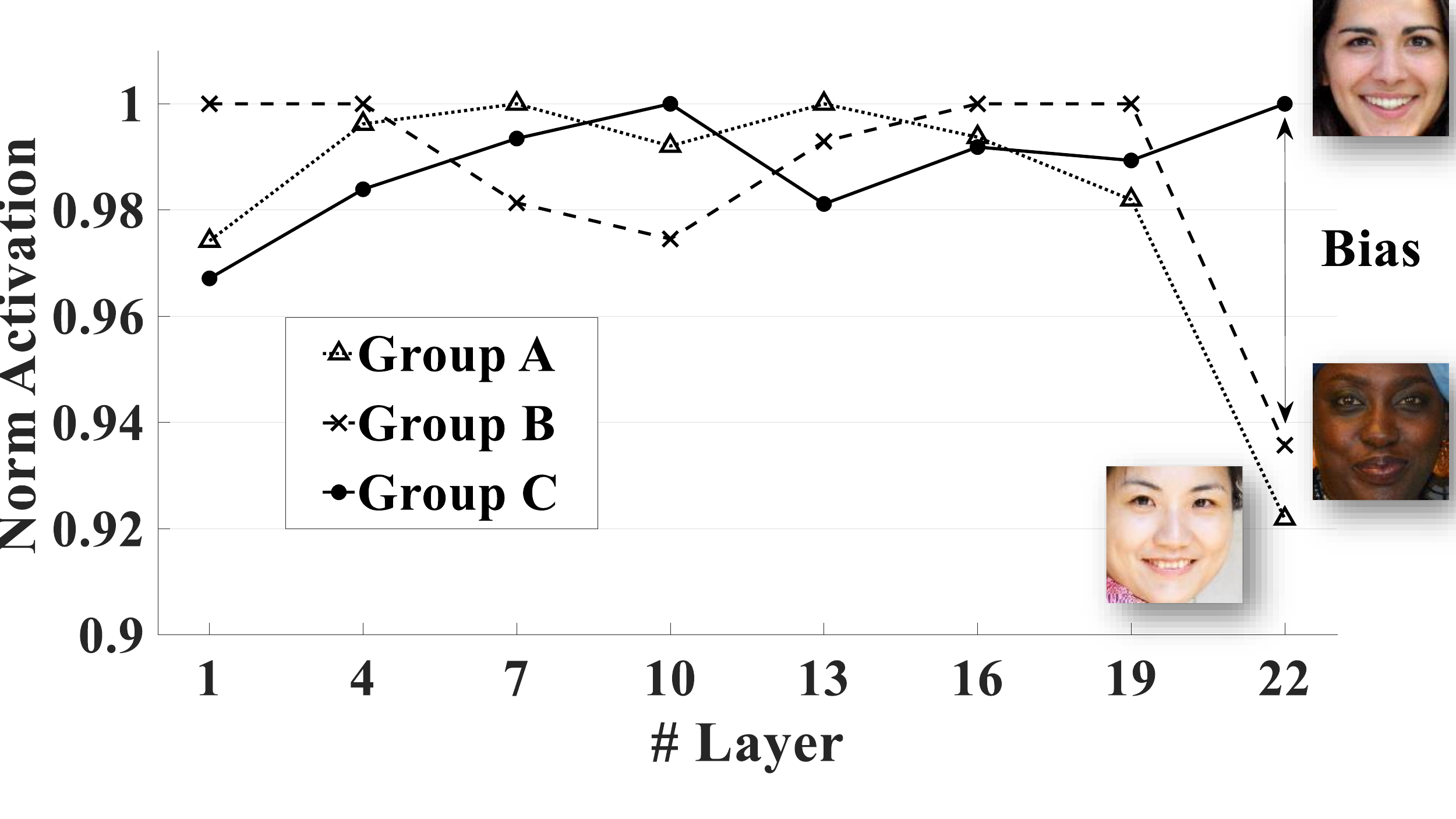}
\caption{Normalized overall activation observed through layers of a biased ResNet model
trained for facial gender recognition. Overall activation for the images of Group A, Group B and Group C (See Section \ref{framework}). The arrow on the right highlights the
difference in activations obtained between well-represented and poorly-represented groups.}
\label{withFaces}
\end{figure}

The number of published works pointing out the biases in the results of 
face recognition algorithms is large \cite{buolamwini2017gender,zisserman2018BlindEye,acien2018bias,nagpal2019Face}.
Among these works, a vast majority analyzes how biases affect the performances obtained for 
different demographic groups \cite{cook2019demographic,buolamwini2017gender,phillips2011Bias,otoole2012demographic,klare2012demographic,han2014demographic,acien2018bias,grother2018FRVT,grother2019FRVT,howard2019effect,isabelle2019DemogPairs,chellapa2019face,drozdowski2020bias}.
However, only a limited number of works analyze how biases 
affect the learning process of these algorithms. In this work we 
use feature visualization techniques of deep models to generate
a better understanding of biased learning. We will analyze two 
gender recognition models based on face images and how the 
ethnicity bias affects the learning process.

Given an input, the activations of the different neurons inside a neural network 
define the output of the network. Fig. \ref{withFaces} shows 
the difference in the overall activation of a gender recognition network 
for input face images from three different demographic groups. The figure shows 
the overall activations at different layers for face images of a well-represented demographic 
group (caucasian female in this example), and 
poorly-represented demographic groups (black and asian females in this example). 
Well-represented group refers to a demographic group prioritized during the training phase of the network
with many more samples compared to poorly-represented groups. 
The overall activation is measured for each image in each layer and then 
averaged across images of the same demographic group. Strong activations 
are usually related with highly discriminative features \cite{zeiler2014visualizing}. 
In the figure, for the well-represented group we can observe how the activations of the gender 
classification network are high for the first layers (focused on 
characterizing textures and colors) increasing slightly for the last layers 
(focused on the high level features related to the gender classification 
task). In comparison, for the poorly-represented group the activations 
decrease significantly towards the final layers.
These low activations may be the cause of biased performances 
that lead to unfair discrimination.

The contributions of this work are twofold: 
\begin{itemize}
    \item Comprehensive analysis of biased learning focused on 
    the effects over the activation level of learned features
    in two public databases; demonstrating the latent correlations between 
    bias, activation, and recognition performance.
    \item  We propose InsideBias, a novel bias detection method 
    based on the analysis of the filters activation of deep 
    networks. InsideBias allows to detect biased models
    with very few samples (only 15 in our experiments), 
    which makes it very useful for algorithm auditing and 
    discrimination-aware machine learning.
\end{itemize} 

The rest of the paper is structured as follows: Section \ref{related} 
summarizes related works in this area. Section \ref{measuring} introduces
the formulation of the problem and the method proposed in
this work. Section \ref{framework} presents the experimental framework
including databases, models, and experimental protocols. Section
\ref{experiments} analyzes experimentally how bias affects different models, and presents the results obtained 
by the  proposed bias detection method InsideBias. Finally, 
Section \ref{conclusions} summarizes the main conclusions.

\section{Related Work}\label{related}

\subsection{Demographic biases in biometrics and AI at large}
The study carried out in \cite{buolamwini2018GenderShades} examined three commercial
face detection algorithms. It revealed the discriminatory 
potential of these technologies and attracted the interest 
of the academia and industry, reaching widespread impact even
in mass media. This presence in the general press comes with a
risk: possible misinterpretation due to excessive generalization
from the particular experiments reported in that paper (based 
on three particular face detection systems) to face biometrics
in general, and AI at large.

The existence of undesired demographic bias and algorithmic
discrimination in AI depend heavily on many factors such as:
learning architecture, training strategy, target problem, and 
performance criteria \cite{serna2020discrimination}. The present work, as others recently \cite{drozdowski2020bias},
generates a better understanding of such factors: 1) as input
knowledge for more informed bias analyses, and 2) to generate 
tools to deal with such undesired biases in practical problems.

As a representative example in this line of work examining
demographic biases in AI, Nagpal et al. conducted a series 
of experiments to verify if deep neural networks in facial 
recognition systems encode some type of race-specific information
\cite{nagpal2019Face},
and found that in models trained with different races, 
different discriminative regions contribute to the decision.

The same algorithm can behave very differently when tested on different groups of samples. For example, \cite{cook2019demographic} demonstrated that same face recognition algorithm performed much worse
for dark-skinned people than for light-skinned people, given images from some cameras, but performed better for images from other cameras.

Turning to the specific case of face recognition, the work \cite{howard2019effect} 
proves that bias is not linked exclusively to demographic factors.
Moreover, it discusses how sensational headlines written by 
non-expert people skew the information around biases in AI. 


\subsection{Looking inside Neural Networks}

As soon as the rebirth of Neural Networks happened in the past decade,
researchers have tried to generate a better understanding of the representations 
learned by neural models.

Erhan et al. proposed an approach to visualize
the hidden layers \cite{erhan2009visualizing}. Zeiler and Fergus applied a
deconvolution algorithm to see the activity within the model 
\cite{zeiler2014visualizing}, and Simonyan et al. generated the representative 
image of a class by maximizing the class scores \cite{simonyan2013visualizing}. 
Yosinski et al. visualized the activations of each neuron when processing an image \cite{yosinski2015visualization}. 
Selvaraju et al. introduced an algorithm that visually highlights a 
network’s decision by computing the gradient of the class
score with respect to the input image \cite{selvaraju2017grad}. 
In a similar line of work, Nguyen et al. created synthetic images that 
maximally activated each neuron \cite{nguyen2016visualization}, 
and Olah et al. explored which neurons are activated in different regions
of the image \cite{olah2018interpretability,olah2017visualization}. 

In addition, there are other variants and improvements 
of the methods indicated above, like the ones that identify 
characteristics encoded by the network relevant to the task 
\cite{oramas2019visual}, or other ways to intervene certain 
neurons in order to see the effect they have \cite{zhou2018neurons}.

Inspired by the literature, in the present work we look at 
the raw activations of the neurons in presence of demographic
biases in gender classification algorithms.

\section{Measuring Bias in Deep Networks: InsideBias} \label{measuring}

\subsection{Formulation of the problem}\label{formulation}

Let’s begin with notation and preliminary definitions. Assume 
$\textbf{I}$ is an input sample (e.g. face image) of an individual.
That sample $\textbf{I}$ is assumed to be useful for task $T$, e.g., face authentication
or gender recognition. That sample is part of a given dataset 
$\mathcal{D}$ (collection of multiple samples from multiple subjects) used to train
a model defined by its parameters $\textbf{w}$. We also assume that there is
a goodness criterion $G_T$ on that task $T$ maximizing some performance function $f_T$ in the given dataset 
$\mathcal{D}$ in the form:

\begin{equation}
    \label{godness}
    G_T(\mathcal{D})=\max_{\textbf{w}}f_T(\mathcal{D,\textbf{w}})
\end{equation}

On the other hand, the individuals in $\mathcal{D}$ can be classified according 
to two demographic criteria (without loss of generality, 
we can have more criteria): $d=1\equiv \textit{Gender} \in \{\textit{Male},\textit{Female}\}$ 
and $d=2\equiv\textit{Ethnicity}\in\{A,B,C\}$. 
We assume that all classes are well represented in dataset $\mathcal{D}$,
i.e., the number of samples of each class for all criteria in $\mathcal{D}$ is
significant. $\mathcal{D}_{d}^k \subset \mathcal{D}$ represents all the samples
corresponding to class $k$ of demographic criterion $d$. 

In our experiments, the goodness criterion $G_T(\mathcal{D})$ is defined 
as the performance of a gender recognition algorithm ($T$ = 
\textit{Gender Recognition}) on the dataset $\mathcal{D}$. During the experiments, we study 
how the criterion $d=1\equiv\textit{Ethnicity}$ affects the internals of 
an algorithm focused on recognition of a different criterion $d=2\equiv\textit{Gender}$.

\subsection{Bias estimation with InsideBias}

While most of the literature is focused on estimating bias through 
performance between different datasets \cite{drozdowski2020bias}, with
InsideBias we propose a novel approach based on the activation
levels\footnote{We refer to activation level as the output of the activation function of each neuron.}
within the network for different datasets $\mathcal{D}_{d}^k$ from 
different demographic groups.

Convolutional Neural Networks are composed by a large
number of stacked filters. These filters are trained to extract 
the richest information for a predefined task (e.g. digit classification or gender 
recognition). These filters are activated as an input (e.g. an image)
goes through the network. Stronger activations are usually related 
to the detection of highly discriminative features \cite{zeiler2014visualizing}.


Without loss of generality, we present InsideBias for Convolutional
Neural Networks (CNNs). Similar ideas are extensible to
other neural learning architectures. In a convolutional layer
of a CNN, the previous layer’s feature maps are convolved
with the filters (also known as kernels) and put through the
activation function to form the output feature map. The output
$\textbf{A}^{[l]}$ of layer $l$ consists of $m^{[l]}$ feature maps of size $n_1^{[l]} \times n_2^{[l]}$, where
$m^{[l]}$ is the number of filters at layer $l$. The $i^{\textnormal{th}}$ 
feature map in layer $l$ denoted as $\textbf{A}_i^{[l]}$ is computed as: 

\begin{equation}
\label{eqn:feature_map}
    \textbf{A}_i^{[l]} = g^{[l]}\left(\sum_{j=1}^{m^{[l-1]}} \textbf{f}_{ij}^{[l]} * \textbf{A}_j^{[l-1]} + \textbf{b}_i^{[l]}\right)
\end{equation}

\noindent where $g^{[l]}$ denotes the activation function of the $l^{\textnormal{th}}$ layer, $*$ is the
convolutional operator, $\textbf{b}^{[l]}_i$ is a bias vector at layer $l$ for the
$i^{\textnormal{th}}$ feature map, and $\textbf{f}_{ij}^{[l]}$ is the filter connecting the $j^{\textnormal{th}}$ feature
map in layer $(l-1)$ with $i^{\textnormal{th}}$ feature map in layer $l$. The average activation of the  $i^{\textnormal{th}}$ feature map at layer $l$ is calculated as:

\begin{equation}
\label{eqn:average}
    \overline{A_i^{[l]}} = \frac{1}{n_1^{[l]} \cdot n_2^{[l]}} \sum_{x=1}^{n_1^{[l]}} \sum_{y=1}^{n_2^{[l]}} A_{i}^{[l]} (x,y)
\end{equation}

\noindent where $(x,y)$ are the spatial coordinates of the output $A_{i}^{[l]}$. The activation, $\lambda^{[l]}$, is calculated as the maximum of $\overline{A_i^{[l]}}$ for all feature maps in the layer $l$: 

\begin{equation}
\label{eqn:activation_prime}
    \lambda^{[l]} = \max_{i}\left(\overline{A_i^{[l]}}\right)
\end{equation}

We have evaluated both the average and the maximum, but the maximum resulted in a better estimator. 
Our intuition is that the maximum is related to highly discriminant patterns (high activations) and this 
is highly correlated to bias effects. 

Filters tend to be different between networks trained differently,
even if the networks have the same architecture, and even if they 
have been trained with the same data. The reason for this is that
if the initialization to solve Eq. \ref{godness} (which is done
iteratively) is different, since the solution space is very large
\cite{lecun2015deep}, the solution of that equation will typically be a
local minimum and will also depend on the particular training
configuration. So to be able to compare the activations of different models we propose to normalize them:

\begin{equation}
\label{eqn:activation}
    \lambda'^{[l]} = \frac{\lambda^{[l]}}{\max_{l}\lambda^{[l]}}
\end{equation}

The \emph{Activation Ratio} $\Lambda^{[l]}_d$ for demographic criterion $d$ (e.g., \textit{Ethnicity} in our experiments) is then calculated as the ratio between 
the activation obtained for the group with the lowest $\lambda^{[l]}$ and 
the group with the highest $\lambda^{[l]}$:

\begin{equation}
    \label{eqn:ratio}
    \Lambda_d^{[l]}= \frac{\min_k\lambda^{[l]}(\mathcal{D}_d^k)}{\max_k\lambda^{[l]}(\mathcal{D}_d^k)}
\end{equation}

InsideBias uses this \emph{Activation Ratio}
to detect biased models. A model will be considered biased 
if the \emph{Activation Ratio} is smaller than 
a threshold $\tau$. When analyzing the bias in this way we
recommend looking at the final layers, similar to the initial example in Fig. \ref{withFaces}.

\section{Experimental Framework}\label{framework}

We start our bias experiments by studying how bias influence the data-driven learning process. For that we trained different architectures for two tasks: digit recognition and gender classification. To better understand its effects and the relationship with the activations, we will analyze the results of training with and without biased training data.

\subsection{Databases}

We have evaluated our approach on two different datasets:

\subsubsection{Colored MNIST}
Inspired in the experiment proposed in \cite{kim2019learning}, we introduced bias in the form of colors into the MNIST dataset \cite{mnist98}. We used the 3 RGB colors (Red, Green, and Blue). Each digit in the training set (60K samples) was colored according to a highly biased distribution (i.e. 90\% of the samples colored with a primary color and 10\% with the remaining two colors). The digits in the test set (10K samples) were colored with an uniform distribution (i.e. 33-33-33). The goal is to analyze the impact of the color information in the learning process of the digit recognition model.

\subsubsection{DiveFace}
The second database used is the DiveFace dataset \cite{SesitiveNets2019}.
DiveFace contains annotations equally distributed among 
six classes related to gender and ethnicity. There are 24K 
identities (4K per class) and 3 images per identity for a 
total number of images equal to 72K. Users are grouped 
according to their gender (male or female) and three 
categories related with ethnic physical characteristics:

\begin{itemize}
\item \textbf{Group A}: people with ancestral origin in Japan,
China, Korea, and other countries in that region.
\item \textbf{Group B}: people with ancestral origins in 
Sub-Saharan Africa, India, Bangladesh, Bhutan, among others.
\item \textbf{Group C}: people with ancestral origins in Europe, 
North-America, and Latin-America (with European origin).
\end{itemize}

\begin{figure}[!t]
\centering
\includegraphics[width=88mm]{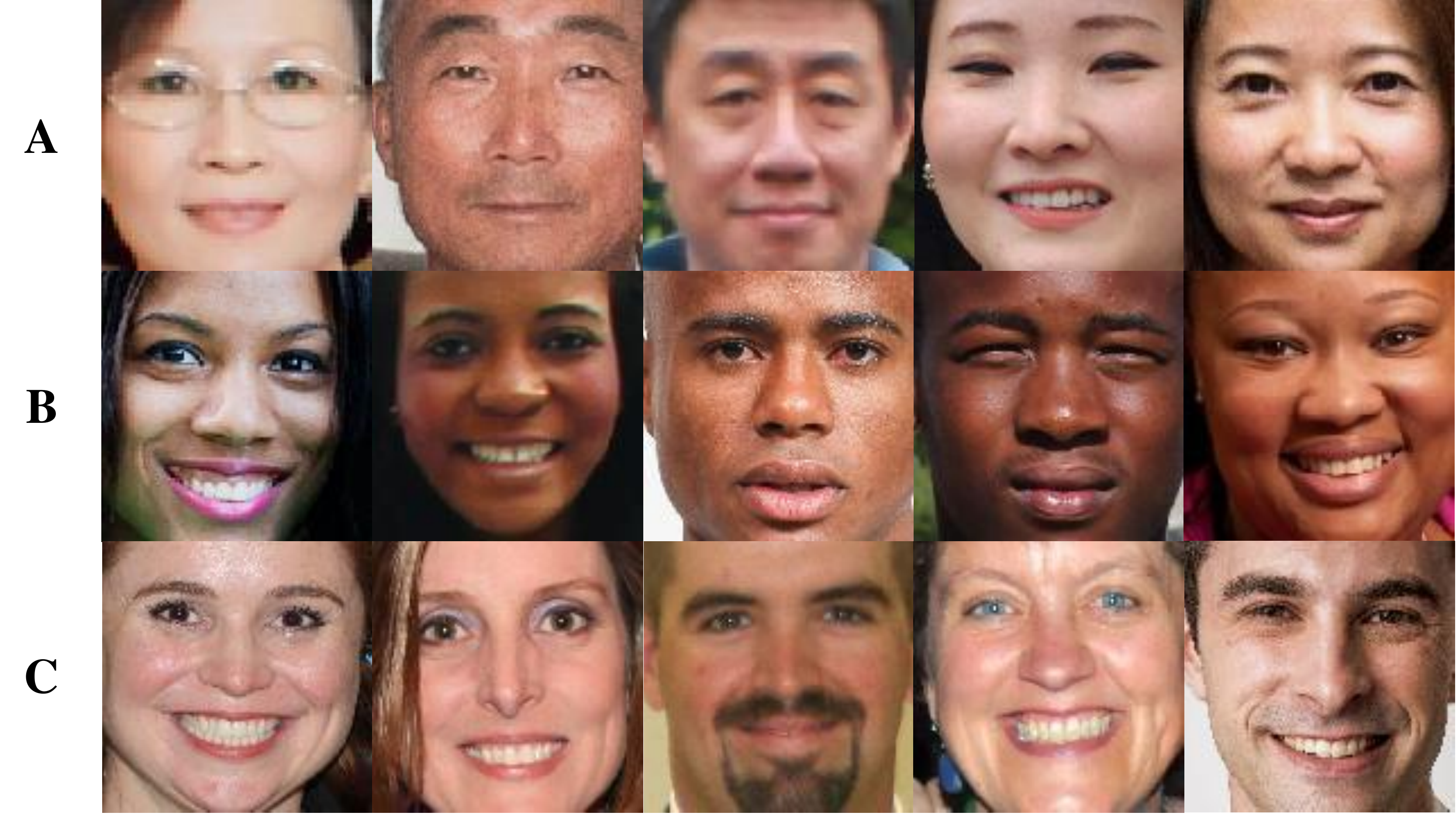}
\caption{Example images for the demographic \textbf{Groups} (\textbf{A, B} and \textbf{C}) used in our experiments.}
\label{images}
\end{figure}

Fig. \ref{images} shows 15 face images examples from the three
ethnic groups of DiveFace. Note that all images show similar 
pose, illumination, and quality. These images obtain very 
high confidence values in the gender recognition algorithms 
of this paper (i.e. confidence scores greater than 99\%). The 
confidence score is the output of the network, it indicates the 
probability of belonging to one class (in our case the class of
being a man or a woman).

Note that these are heterogeneous groups that include people
of different ethnicities. We are aware of the limitations of
grouping all human ethnic origins into only three categories.
According to studies, there are more than 5K ethnic groups 
in the world. We have classified them into only three groups
in order to maximize differences between classes. Automatic 
classification algorithms based on these three categories 
show performances of up to 98\% accuracy \cite{acien2018bias}.

\subsection{Learning architectures}\label{architect}

We employed two popular state-of-the-art image recognition
architectures based on Convolutional and Residual layers. 
These architectures have been chosen as examples of standard
models employed in face attribute detection algorithms \cite{ranjan2018faces,sosa2018SoftBiometrics}:

\textit{Network 1 (VGG architecture \cite{parkhi2015face}):} The network is composed 
of eight convolutional layers followed by two fully connected 
layers with dropout. We use the ReLU (Rectified Linear Unit) 
activation function in all hidden layers, and a softmax 
activation function for the output layer (with two output units). 
This network comprises more than $660$K
parameters and its input is $120\times120$ for the gender recognition model and $28\times28$ for the digit recognition model. 

\textit{Network 2 (ResNet architecture \cite{he2016resnet}):} The network consists 
of three building blocks and a fully connected layer with 
softmax activation for the output layer (with two output units).
Each building block is composed of convolutional layers. The big difference with the VGG architecture is the shortcut connections: within each block there is a shortcut connection that performs a convolution and bypasses a certain number of convolutional layers.
This network comprises more than $370$K parameters and its input is $120\times120$.

\subsection{Experimental protocol}\label{protocol}

\subsubsection{Colored MNIST Protocol} We have trained 31 digit recognition models using the VGG architecture: 

\begin{itemize}
    \item  \textit{Biased Models:} to analyze the impact of biased training data on the learning process, we decided to apply a highly biased color distribution in each of the ten digits of MNIST dataset. We defined three possible colors for a digit (Red, Green, and Blue). The highly biased distribution means that 90\% of the training samples of a digit are colored with a primary color (e.g. Red), and the other 10\% with the secondary colors (e.g. Green and Blue). Of the remaining 9 digits, all their training samples are colored with one of the secondary colors. This process is repeated for the ten digits and the three colors, resulting in 30 different models with 30 different biases.
    \item \textit{Unbiased Model:} one model is trained with uniform color distribution (33\%-33\%-33\%).
\end{itemize}

The color in the test set is assigned uniformly 33\%-33\%-33\%. The goal of this experimental protocol is to analyze the relationship between bias, performance, and activations.   


\subsubsection{Gender Recognition Protocol} We have trained four models of each of the two chosen learning architectures,
according to three different experimental protocols:

\begin{itemize}
    \item  \textit{Biased Models}: the models in this experiment are trained with 18K
    images giving priority to one ethnic group with 90\% of the images as opposed to the other two ethnic groups which are 5\% and 5\% respectively (all divided equally between men and women). This experiment is repeated 3 times giving preference to each
    ethnic group. Therefore, 3 independent models per network architecture are trained.
    \item  \textit{Unbiased Model} (trained with limited data): the models in this experiment are
    trained with 18K images (same number of images than biased models), 6K from each ethnic
    group, divided in half between men and women.
\end{itemize}

All models are evaluated with 18K images distributed equally
among all three ethnic groups. None of the validation 
users have been used for training; i.e., it is an independent
set.

\section{Experimental Results}\label{experiments}

\begin{figure*}[!t]
\centering
\subfloat[Red \textit{Biased} (56.86\%)]{\includegraphics[width=45mm]{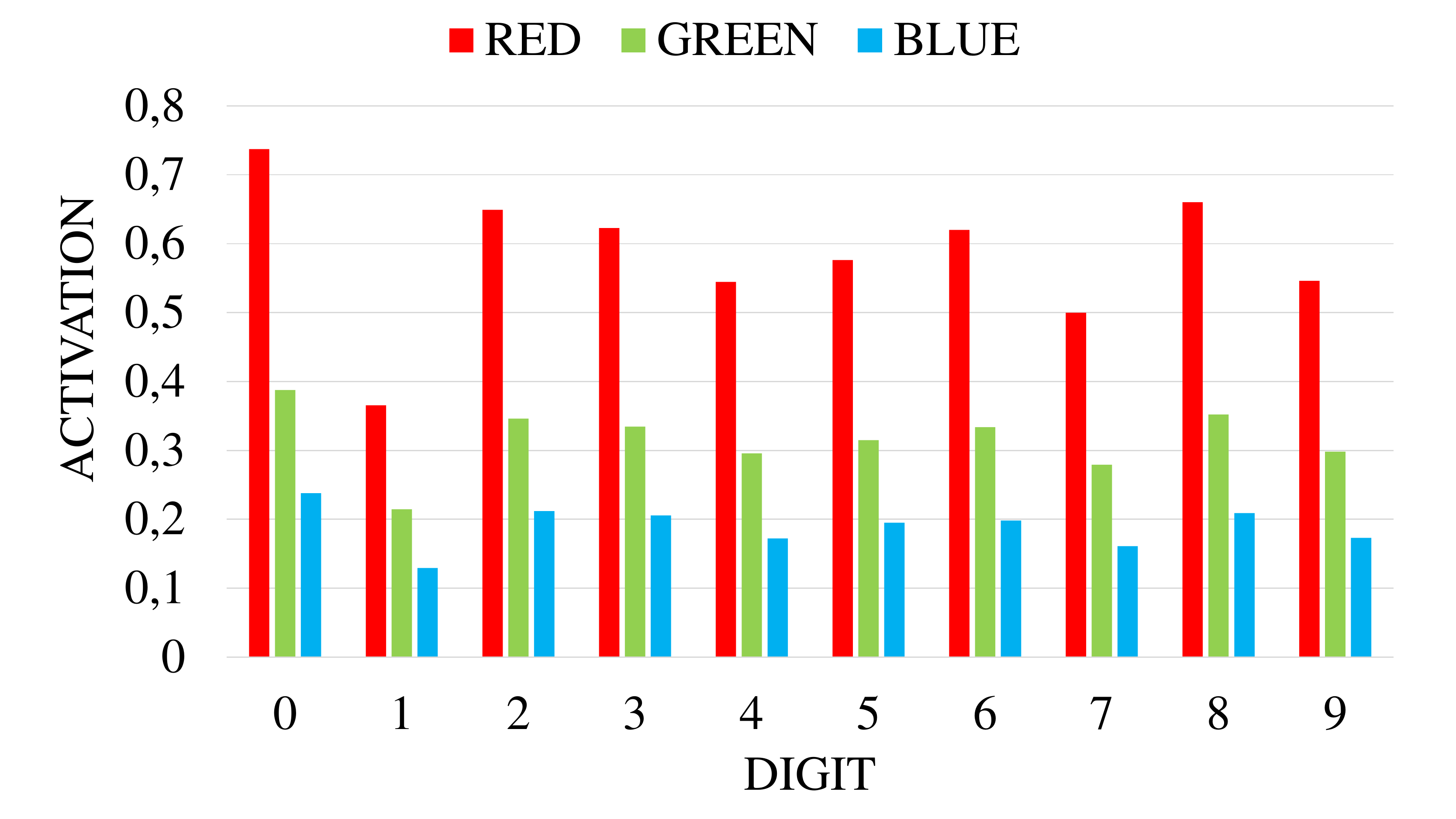}%
\label{Red}}
\hfil
\subfloat[Green \textit{Biased} (56.98\%)]{\includegraphics[width=45mm]{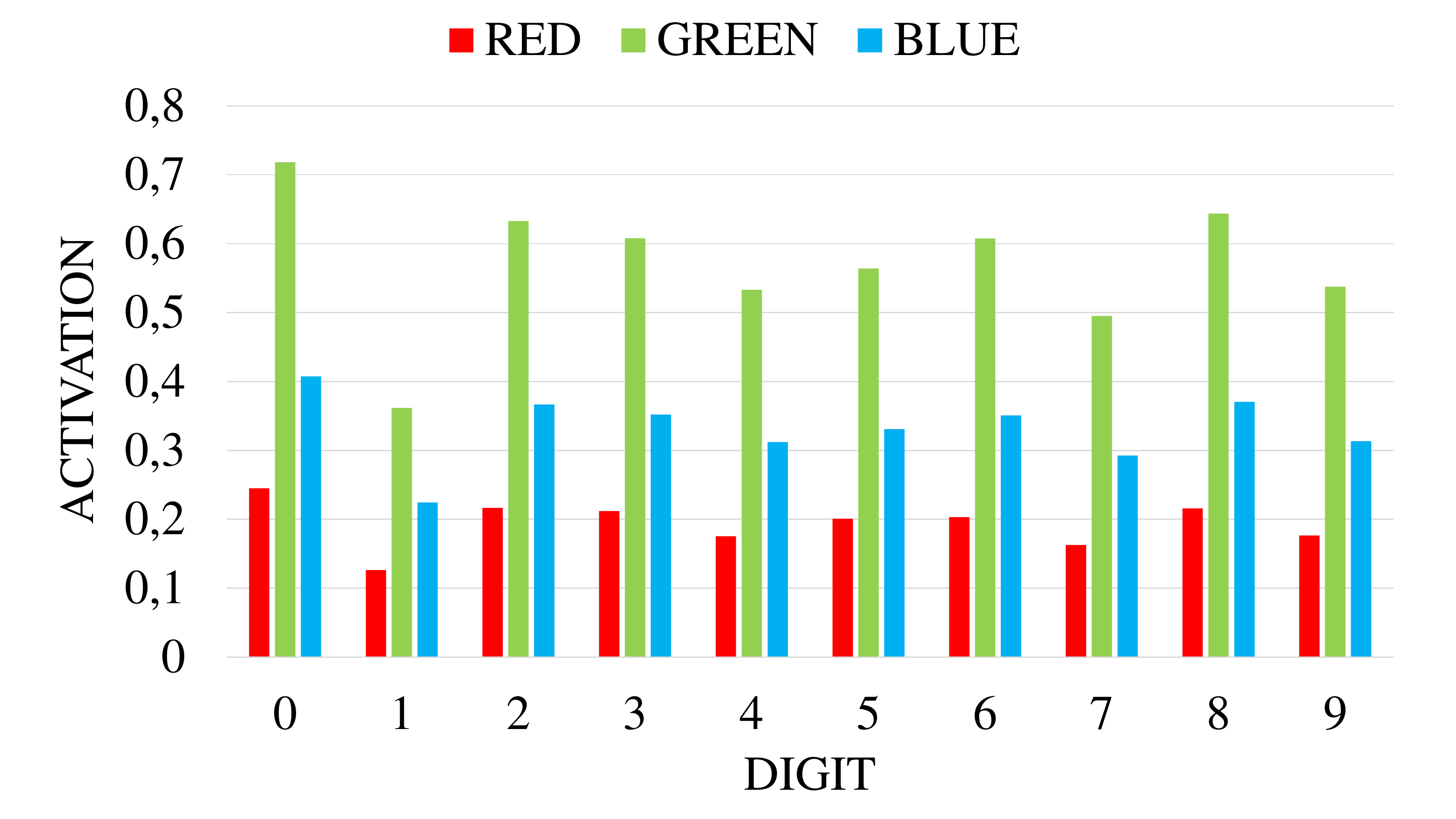}%
\label{Green}}
\hfil
\subfloat[Blue \textit{Biased} (56.35\%)]{\includegraphics[width=45mm]{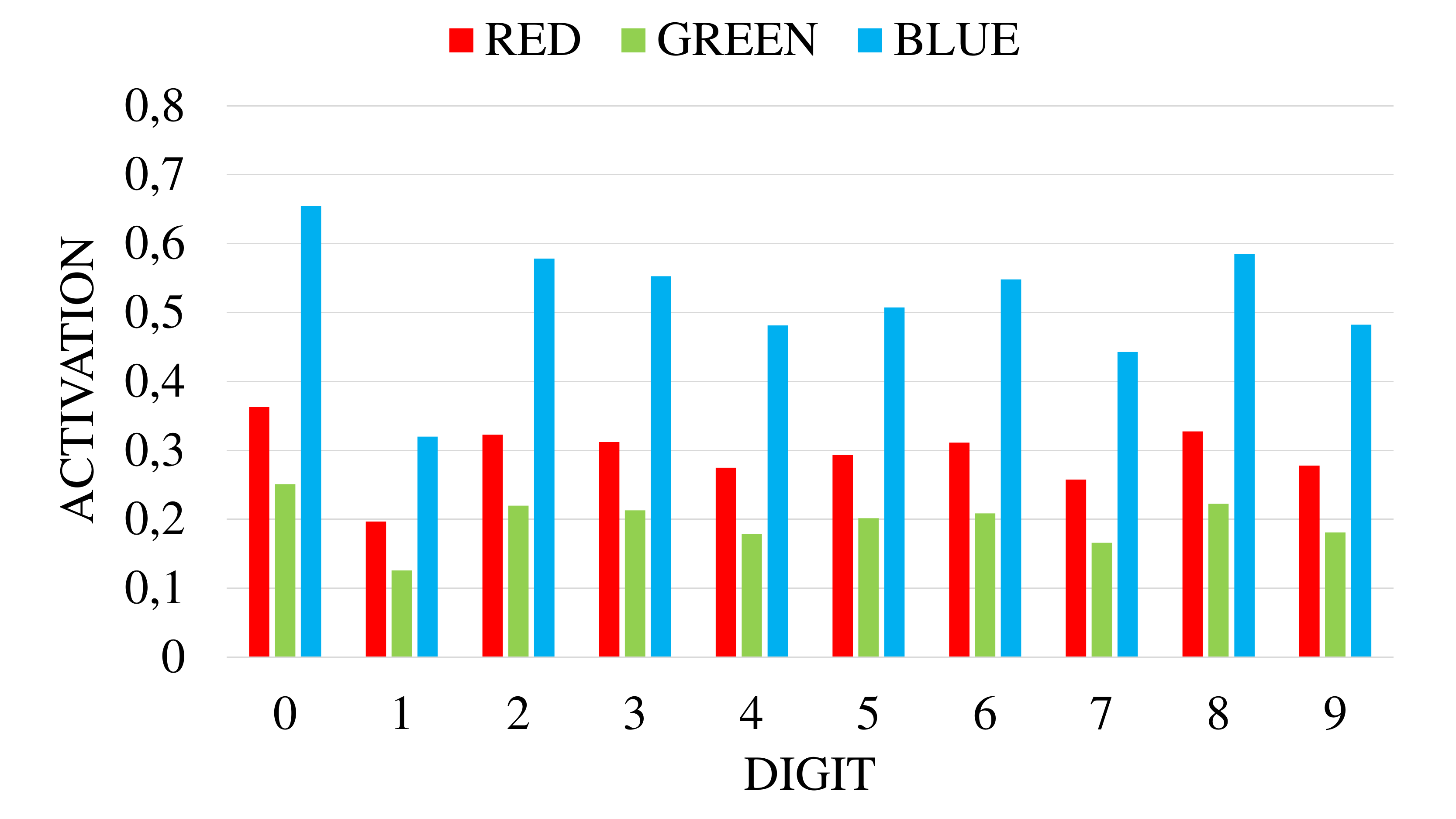}%
\label{Blue}}
\hfil
\subfloat[\textit{Unbiased} (97.87\%)] {\includegraphics[width=45mm]{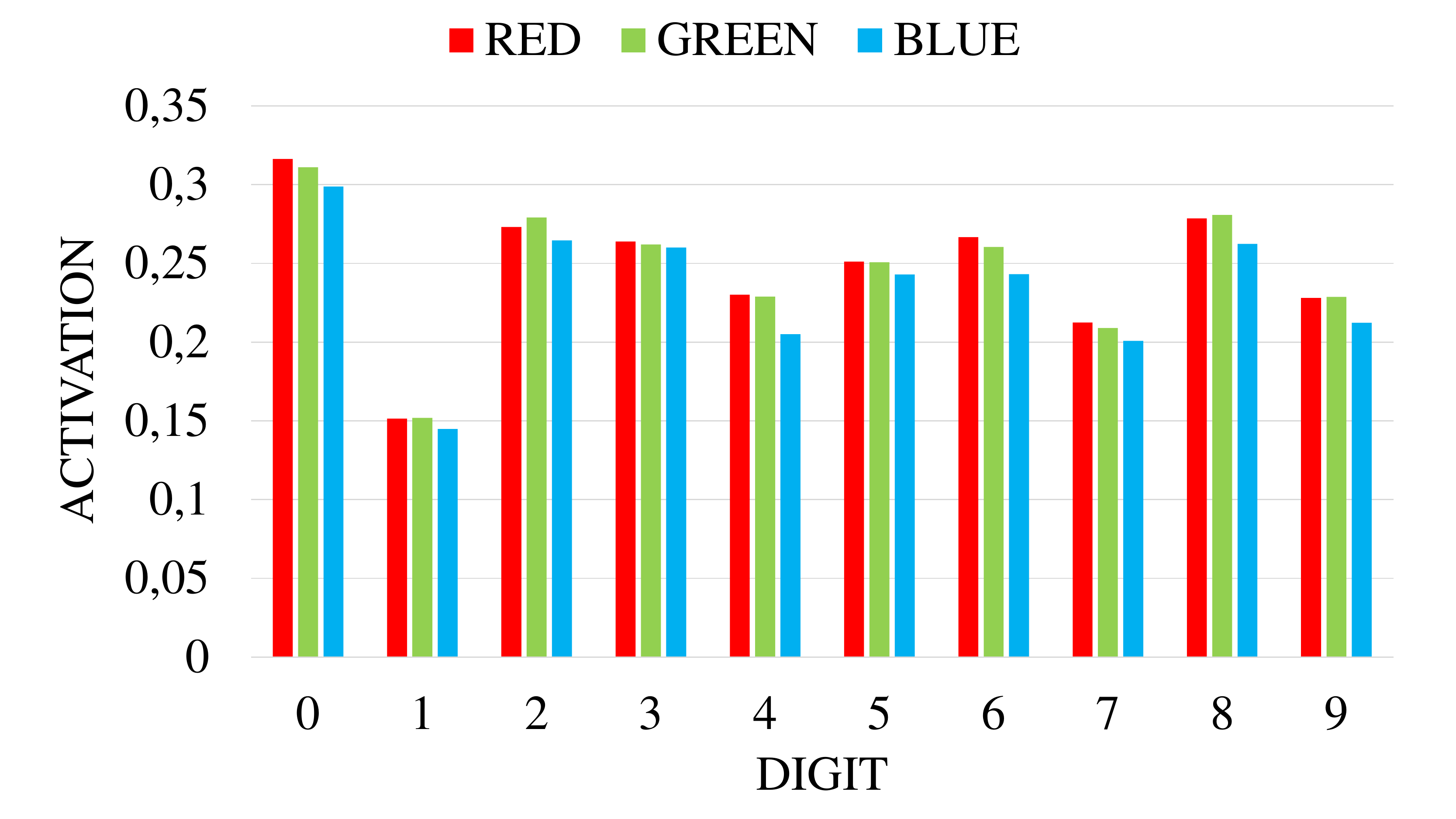}%
\label{Eq}}
\caption{Average $\lambda^{[l]}$ activation observed in the test set (10K samples) for the digits of each color in the last convolutional layer of the models trained in a biased way according to the three color distributions (a-c) and for the unbiased model (d). In brackets the classification accuracy obtained for the test set.}
\label{mnist}
\end{figure*}

\subsection{Role of the biased data}

\subsubsection{Colored MNIST} Fig. \ref{mnist} shows the average activation $\lambda^{[l]}$ for the different models trained using the colored MNIST dataset and the protocol described in Section \ref{protocol}. The results demonstrate the correlation between bias, performance and activations. The poor performance obtained by highly biased models is not surprising. The biased distribution of color introduced in the training set decreases the performance of the network when testing in a set with uniform distribution of color. The experiment with the colored MNIST suggests that bias also affects the activations. On the one hand, a model trained with a color bias tends to produce higher activations for this color (see Fig. \ref{mnist}.a-c). On the other hand, a model trained without bias (i.e. a uniform distribution of colors in the training set) produces similar activations for the different colors (see Fig. \ref{mnist}.d). These results suggest the potential of the proposed $\lambda^{[l]}$ activation as a good estimator of the bias in a trained model.

\subsubsection{Gender Recognition} Table \ref{tabla} shows the results of the experiments described 
in Section \ref{protocol}, performed on \emph{Networks} 1 and 2, 
respectively (see Section \ref{architect}). The results show that the 
models trained using data from a single ethnic group perform 
better for this group (\textit{Biased Models}). These results suggest 
that the ethnic features affect the performance of gender 
recognition based on the two popular network architectures evaluated.
The learned parameters $\textbf{w}$ of a model 
trained with one ethnic group (for example ethnic group $k=1 \equiv \textit{Asian}$) 
do not generalize in the 
best possible way for other groups, with a clear drop in 
performance between testing groups for both networks, i.e., 
using the notation introduced in Section \ref{formulation}: leaving fixed the network trained for $T=\textit{Gender classification}$ by maximizing the
goodness criterion $G_T$ over $\mathcal{D}_{\textit{Ethnicity}}^{\textit{Asian}}$ (see Equation \ref{godness}), we observe that
$G_{T}(\mathcal{D}_{\textit{Ethnicity}}^{\textit{Asian}}) >> G_T(\mathcal{D}_{\textit{Ethnicity}}^{\textit{African}})$ and $G_T(\mathcal{D}_{\textit{Ethnicity}}^{\textit{Caucasian}})$.

On the other hand, training another network also for $\textit{Gender classification}$ in this case with unbiased data representing well all ethnic groups (\textit{Unbiased 
Models}) and leaving it fixed, reduces the performance gap between testing groups 
and improves the overall accuracy (Avg in Table \ref{tabla}) i.e. $G_\textit{Gender}(\mathcal{D}_{\textit{Ethnicity}}^\textit{Asian}) \approx G_\textit{Gender}(\mathcal{D}_{\textit{Ethnicity}}^{\textit{African}}) \approx G_\textit{Gender}(\mathcal{D}_{\textit{Ethnicity}}^{\textit{Caucasian}})$.
However, the performance achieved by the \textit{Unbiased Models} trained with 
heterogeneous data does not improve the 
best performance achieved by each of the \textit{Biased Models}
trained using data only from one ethnic group. 

\begin{table}[!t]
    \normalsize
    \caption{Accuracy (\%) in gender classification for VGG and ResNet models for each of the three demographic groups. Each line indicates a model. The Protocol column indicates the ethnic group employed to train, the Group columns indicate the testing groups, and the other columns: average Accuracy across groups (Avg) and standard deviation (Std) (lower means fairer)}\smallskip 
    \centering
    \begin{tabular}{l c c c c c}
      \toprule
      \multicolumn{6}{c}{VGG}\\
      \midrule 
      \textbf{Protocol} & \textbf{A} & \textbf{B} & \textbf{C} & \textbf{Avg} & \textbf{Std}  \\
      \midrule 
        Biased (\textit{A})    & \textbf{95.72} &   94.16   &	94.68 &	94.85 &	0.65 \\
        Biased (\textit{B})    & 94.16    &	\textbf{95.82}   &	94.06 &	94.16 &	1.31  \\
        Biased (\textit{C})    & 92.46    &	94.63   &	\textbf{96.71} &	94.60 &	1.74 \\
        Unbiased  & 94.84    &	95.69   &	95.28 &	95.27 &	0.34 \\
      \bottomrule 
      \toprule[0pt]
      \multicolumn{6}{c}{ResNet}\\
      \midrule 
      \textbf{Protocol} & \textbf{A} & \textbf{B} & \textbf{C} & \textbf{Avg} & \textbf{Std}\\
      \midrule 
        Biased (\textit{A})    & \textbf{96.84} &   94.14   &	94.45 &	95.14 &	1.21 \\
        Biased (\textit{B})    & 93.29  &	\textbf{96.86}   &	95.40 &	95.18 &	1.47  	\\
        Biased (\textit{C})    & 94.80  &	95.21   &	\textbf{97.01} &	95.67 &	0.96 \\
        Unbiased  & 95.50  &	95.35   &	96.11 &	95.65 &	0.33  \\
      \bottomrule 
    \end{tabular}
\label{tabla}
\end{table}

\begin{figure*}[!t]
\centering
\subfloat[VGG \textit{Biased} (\textit{A)}]{\includegraphics[width=44mm]{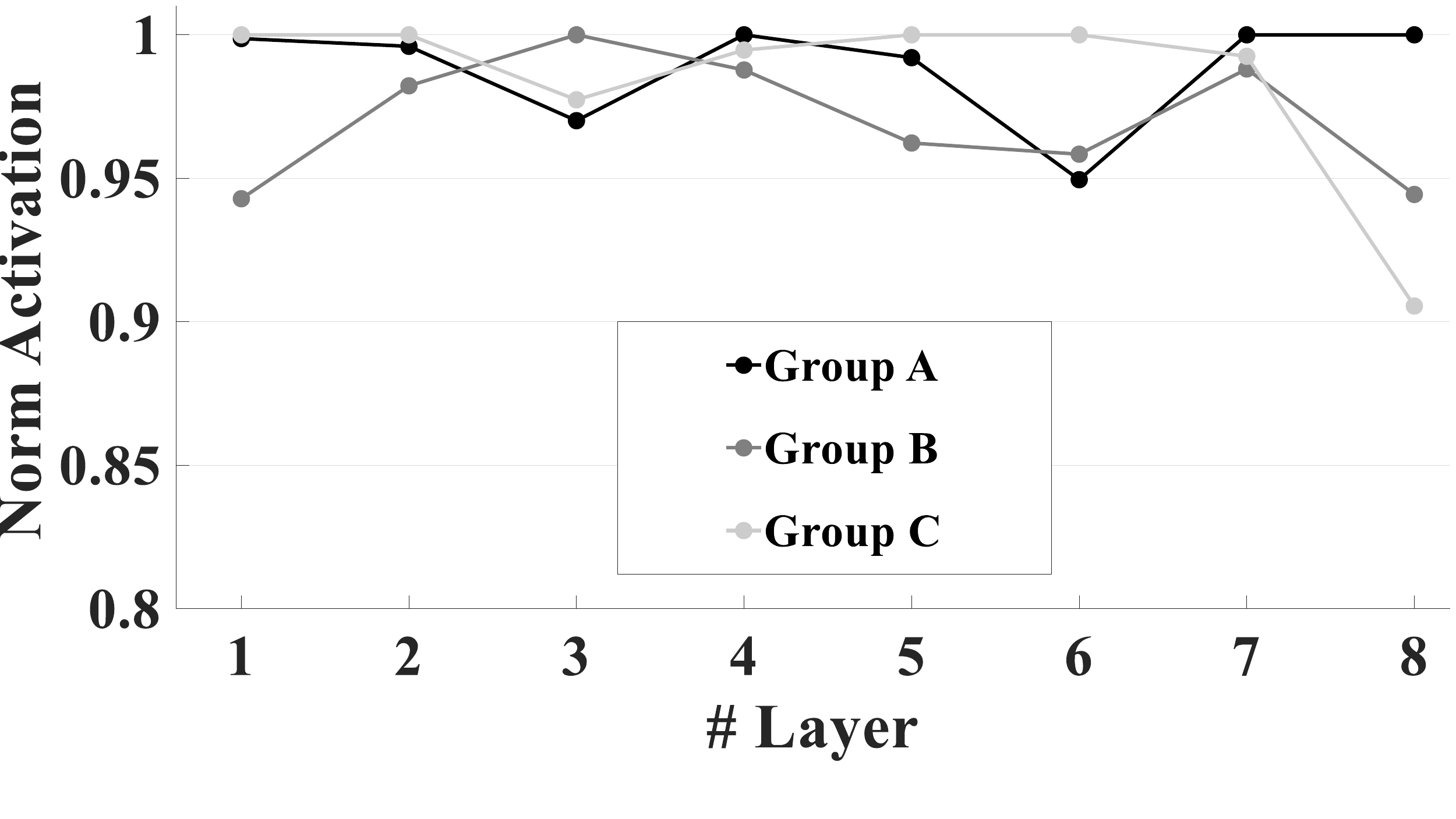}%
\label{VGG_A}}
\hfil
\subfloat[VGG \textit{Biased} (\textit{B)}]{\includegraphics[width=44mm]{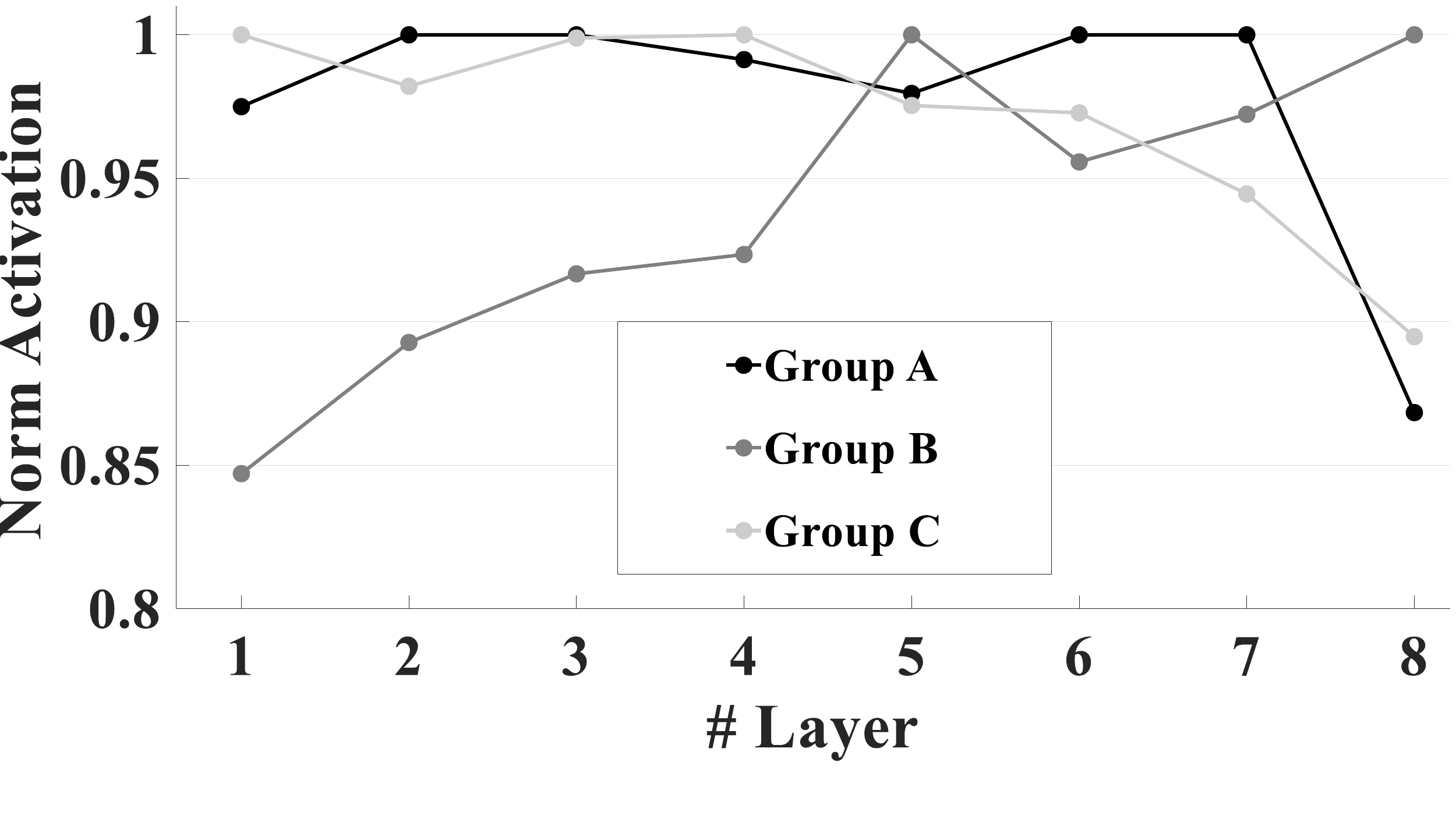}%
\label{VGG_B}}
\hfil
\subfloat[VGG \textit{Biased} (\textit{C)}]{\includegraphics[width=44mm]{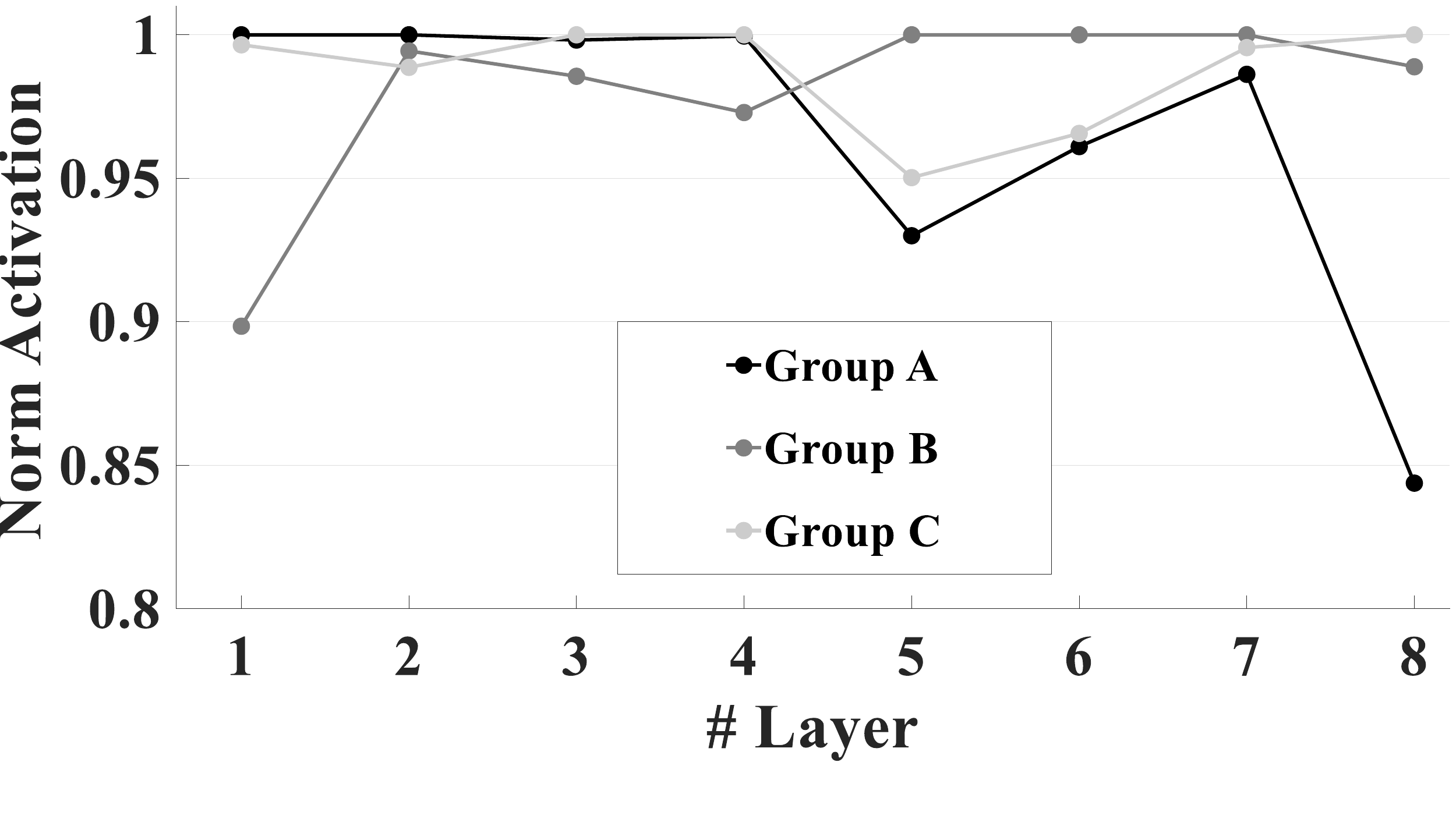}%
\label{VGG_C}}
\hfil
\subfloat[VGG \textit{Unbiased}] {\includegraphics[width=44mm]{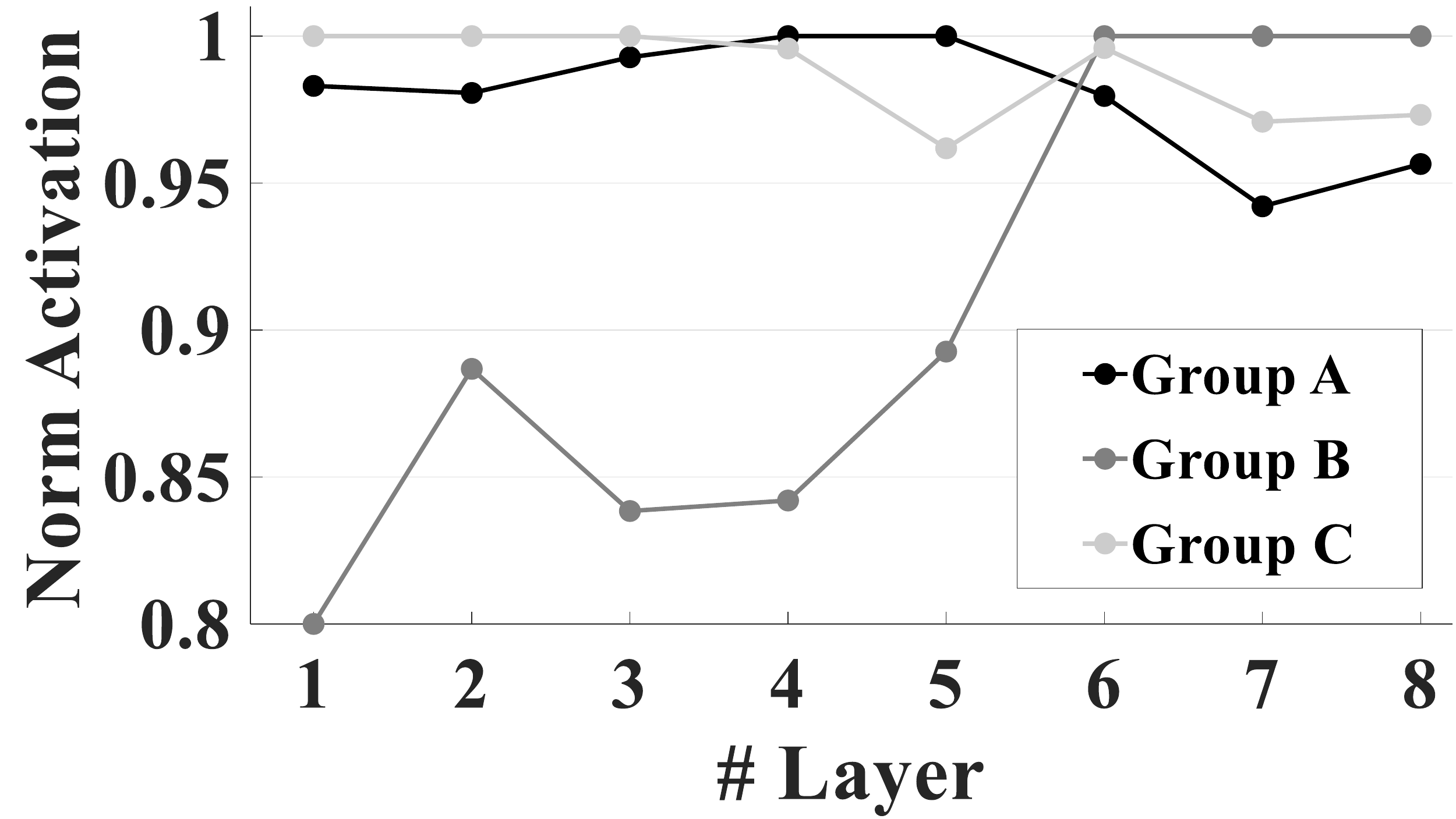}%
\label{VGG_All}}
\hfil
\subfloat[ResNet \textit{Biased} (\textit{A)}] {\includegraphics[width=44mm]{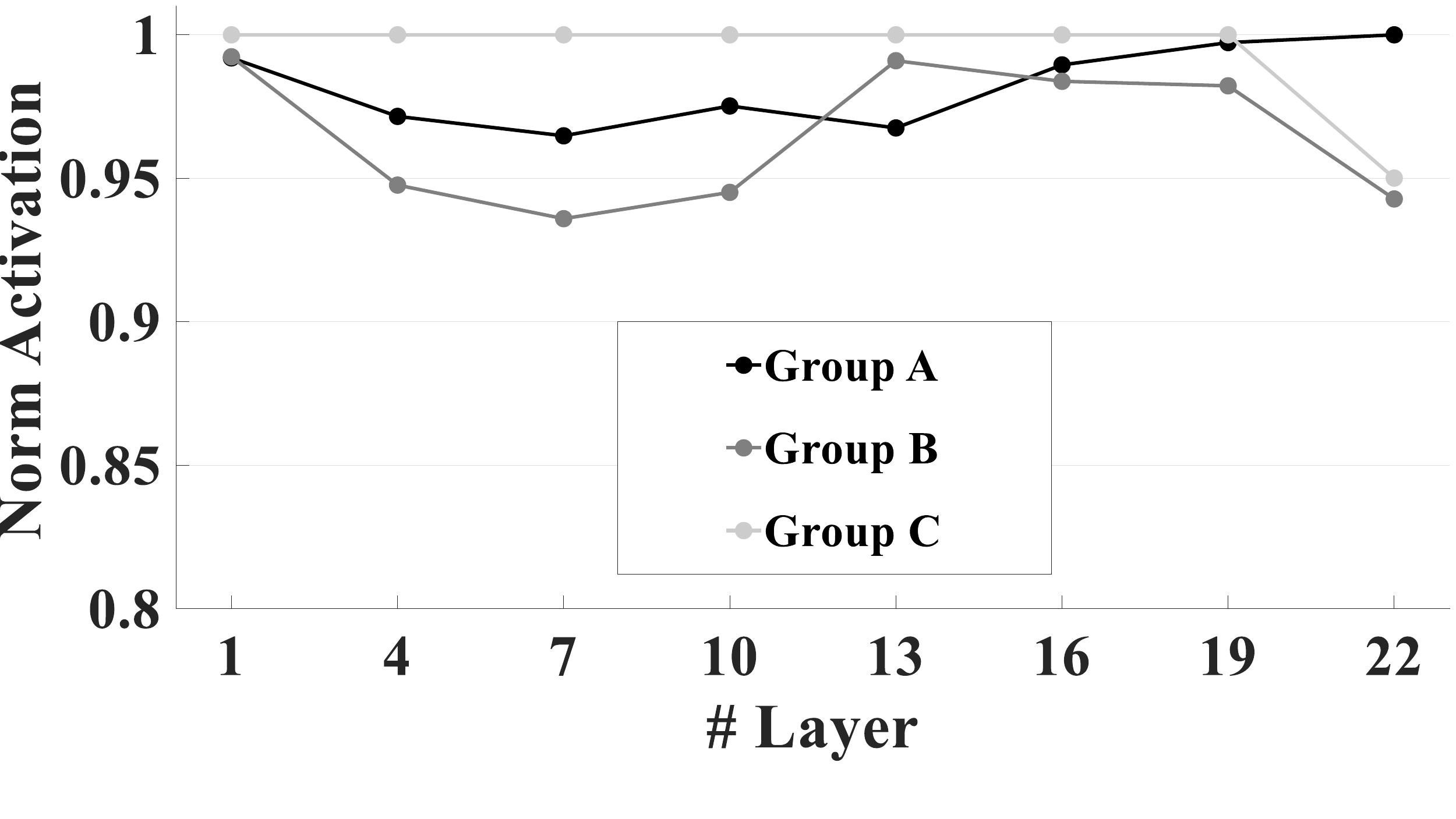}%
\label{ResNetA}}
\hfil
\subfloat[ResNet \textit{Biased} (\textit{B)}] {\includegraphics[width=44mm]{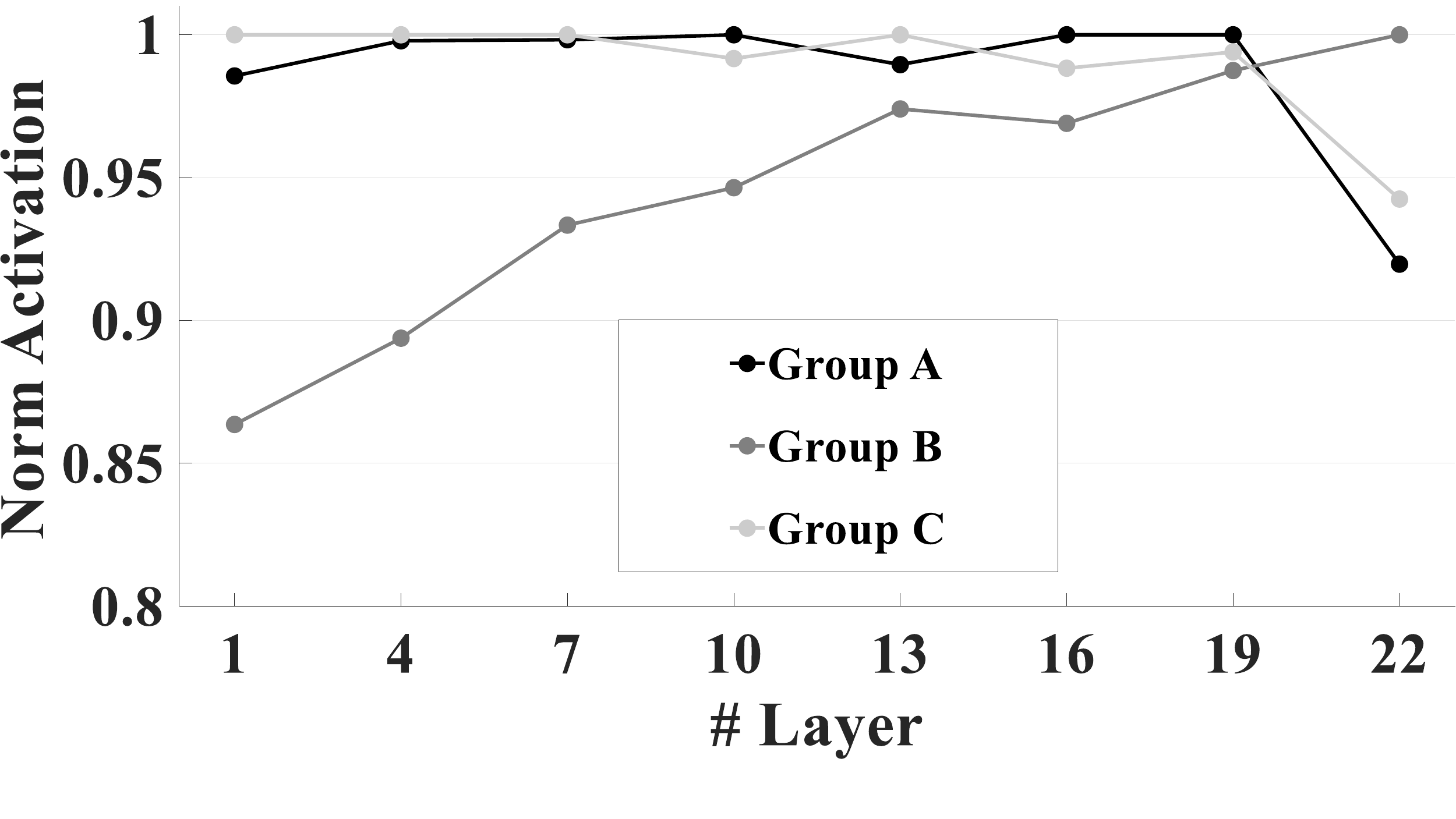}%
\label{ResNetB}}
\hfil
\subfloat[ResNet \textit{Biased} (\textit{C)}] {\includegraphics[width=44mm]{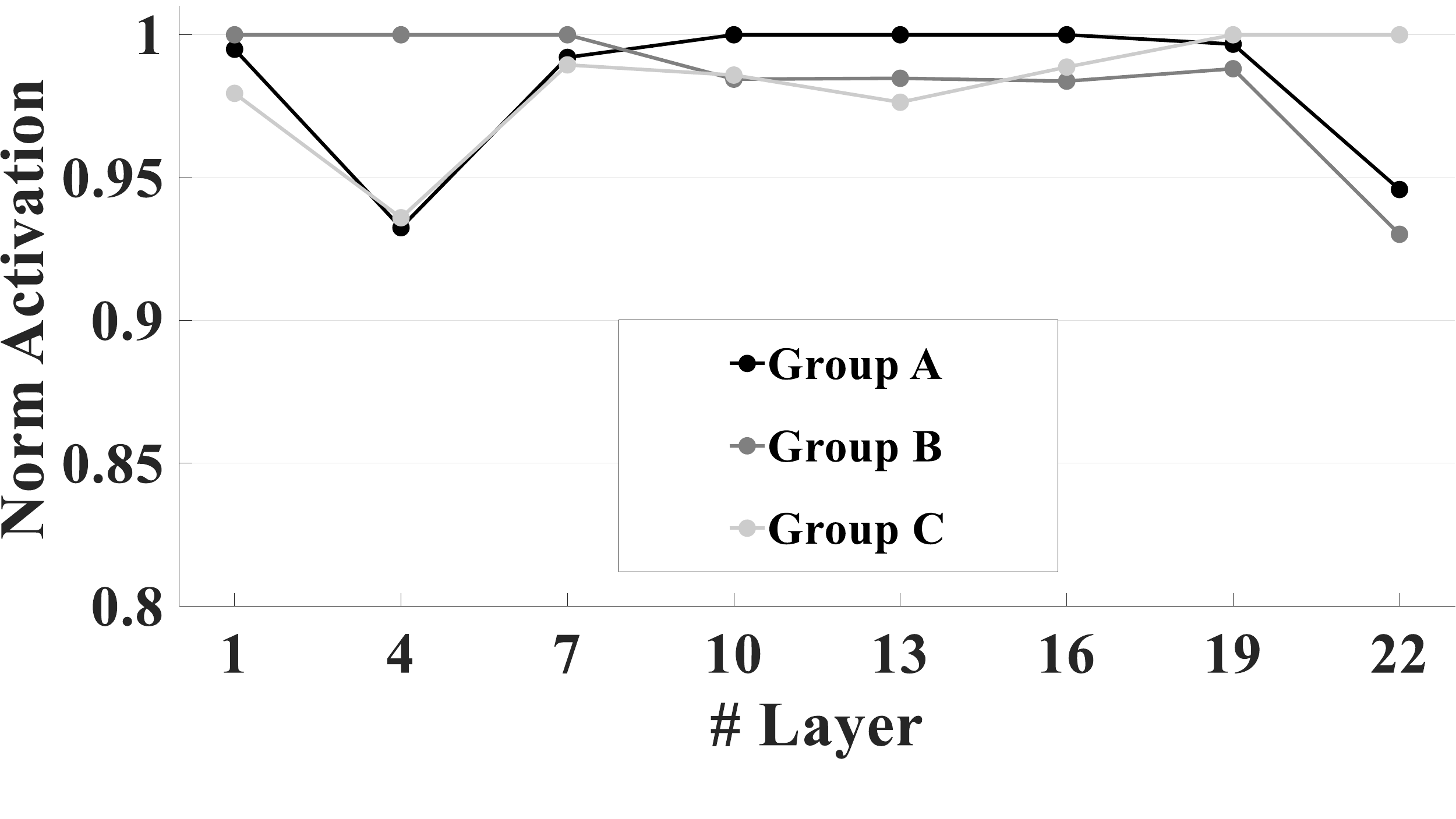}%
\label{ResNetC}}
\hfil
\subfloat[ResNet \textit{Unbiased}] {\includegraphics[width=44mm]{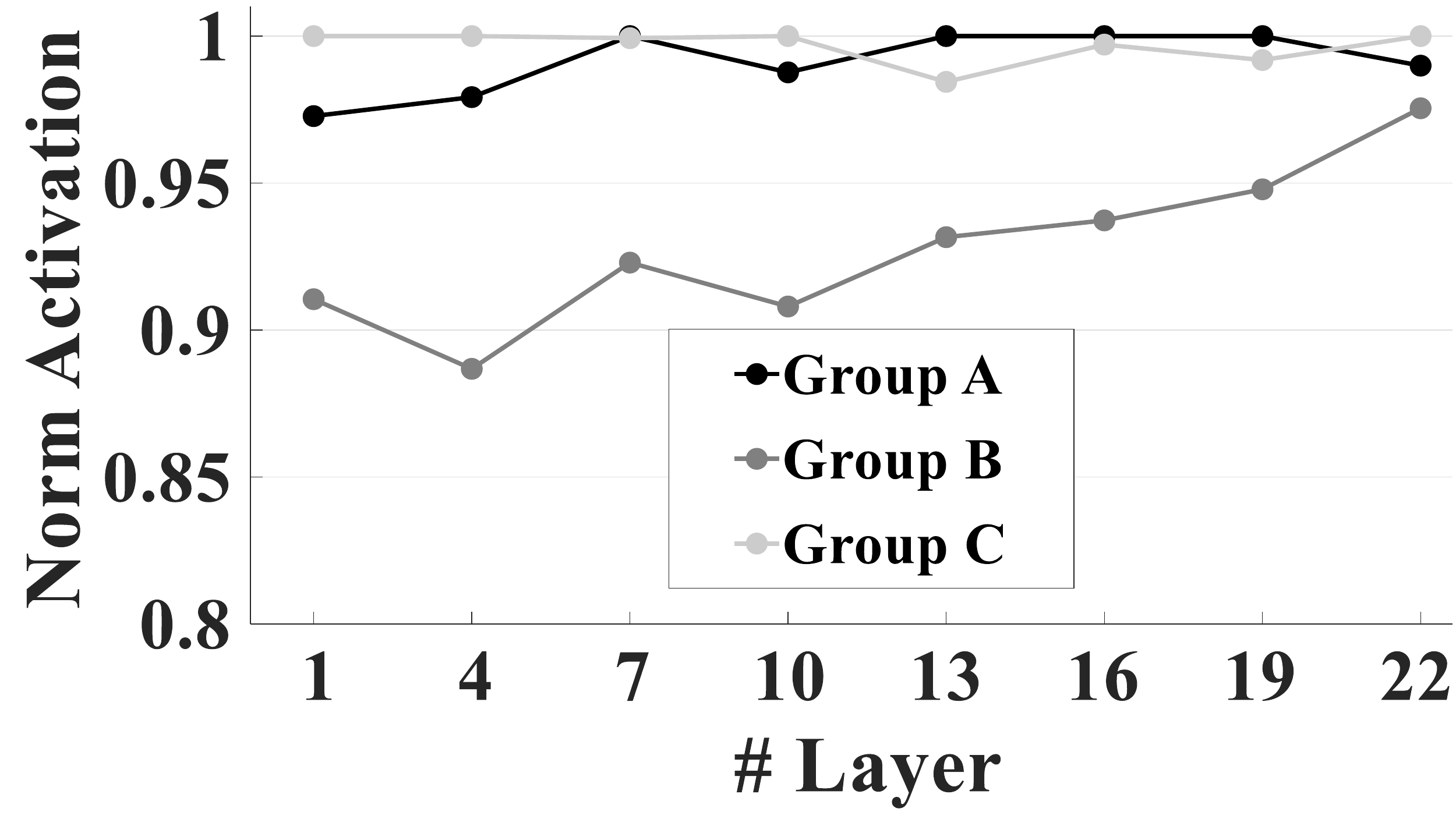}%
\label{ResNet_All}}
\caption{Normalized activation $\lambda'$ observed in \textbf{testing} ($18$K images) for the three demographic \textbf{Groups} (\textbf{A}, \textbf{B} and \textbf{C}) of the different \textit{trained} models: \textit{Biased Model} \textit{(A}, \textit{B} and \textit{C}) and \textit{Unbiased Model}. The top row shows the activations of \textit{Network} 1 (VGG) and the bottom row plots the activations of \textit{Network} 2 (ResNet). The VGG model has only 8 convolutional layers while the ResNet model has 22 (only the main ones appear).}
\label{fig_sim}
\end{figure*}

\subsection{InsideBias: Activation as a bias estimator}

Fig. \ref{fig_sim} shows the \emph{Normalized Activation} from Equation \ref{eqn:activation} of the different networks for each demographic group.
Fig. \ref{fig_sim} shows that 
the activations obtained for the \textit{Unbiased Models} in the last layers have the lowest differences between ethnic groups in testing. As we can
see in the activation curves by layer, for \textit{Biased
Models} the differences in the activation between well-represented
and poorly-represented groups are not homogeneous across 
layers. The curves suggest greater activation differences between
groups, in this case (\emph{Biased Models}), specially in the 
last layers of the networks.

We also see in Fig. \ref{fig_sim} in the first layers that testing \textbf{Group B} gets considerably less activation than the other two groups
(mainly in the models trained with this group, see \ref{VGG_B} and \ref{ResNetB}), which 
tells us that this group has less activation for layers extracting low level features (e.g. shape, texture, and colors). 
However, this lower activation in the first layers does not necessarily imply a low performance (as seen in Table \ref{tabla}).
The low activation in the first layers is compensated with high activation in the last layers which are related with high level features close to the task $T$ (i.e. gender recognition in our experiments). These results suggest a correlation between the bias introduced in Section \ref{protocol}, the performance reported in Table \ref{tabla}, and the activations showed in Fig. \ref{fig_sim}.

\subsection{InsideBias: Detecting bias with very few samples} \label{Diff}

The activations presented in Fig. \ref{fig_sim} shows the relationship between bias and activations over $18$K images,
$2$ learning architectures, and $5$ models trained according to
different biased and unbiased datasets. The following experiments investigate how InsideBias performs for bias detection using a small set of test samples. 

Table \ref{tabla_activation} shows the average classification scores and the \emph{Activation Ratio} defined in Equation \ref{eqn:ratio} ($l=$ last layer) obtained with the
different considered models (Biased and Unbiased) trained for gender classification and tested with the 15 face images shown in Fig. \ref{images}. The results show how 
activations are correlated with biases even if the classification scores show almost no differences
between biased and unbiased models. An \emph{Activation Ratio} close to one reveals a similar activation pattern for all demographic groups tested, which is what we have for the \emph{Unbiased models}
(i.e. $\lambda^{[l]}(\mathcal{D}_{\textit{Ethnicity}}^{\textit{Asian}}) \approx \lambda^{[l]}(\mathcal{D}_{\textit{Ethnicity}}^{\textit{African}}) \approx \lambda^{[l]}(\mathcal{D}_{\textit{Ethnicity}}^{\textit{Caucasian}})$). 
In contrast, the \emph{Biased models} show lower \emph{Activation Ratio}, which means higher differences between activation patterns from 
well-represented and poorly-represented demographic groups (i.e. $\lambda^{[l]}(\mathcal{D}_{\textit{Ethnicity}}^{\textit{Well-represented}}) > \lambda^{[l]}(\mathcal{D}_{\textit{Ethnicity}}^{\textit{Poorly-represented}}$). 

These results suggest that even if the network was trained 
only for gender recognition, the activation level of the filters is 
highly sensitive to the ethnic attributes. The proposed method 
for bias detection in deep networks, InsideBias, consists 
of measuring that sensitivity with the \emph{Activation Ratio} $\Lambda^{[l]}_d$
defined in Equation \ref{eqn:ratio} and comparing it to a threshold $\tau$.

The main advantage of this method for the analysis of bias 
with respect to a performance-based evaluation is that the 
differences are examined in terms of model behavior. Images 
of Fig. \ref{images} obtained good performance (over 99.99\% confidence
score even in biased models) but showed clearly different activation patterns $\lambda$. Bias analysis based 
on performance require large datasets, and using the proposed
\emph{Activation Ratio}, few images may be enough to detect
biased models. 

In this work we do not underestimate the performance as
a good instrument for analyzing bias in deep networks. We
propose to include activation as an additional evidence \cite{fierrez2018Fusion2}, 
which is specially useful when only very few samples are 
available for bias analysis.

\begin{table}[!t]
    \normalsize
    \caption{Average gender confidence scores $S$ and Activation Ratios $\Lambda^{[l]}_d$ obtained by the biased an unbiased models tested for the $15$ images of Fig. \ref{images}. $l=$ last convolutional layer. $1$ is 100\% confidence about the true gender attribute in the image.}\smallskip 
    \begin{center}
    \begin{tabular}{l|l|c c c c}
      \toprule
      \multicolumn{2}{}{} & \multicolumn{3}{c}{\textbf{Test Group}} & \multirow{2}{*}{$\Lambda^{[l]}_d$}\\
      \cmidrule{3-5}
      \multicolumn{2}{l}{\textbf{Model} (\textbf{\textit{Training}})} & \textbf{A} & \textbf{B} & \textbf{C} &\\
      \midrule 
       \multirow{2}{*}{VGG-Biased (\textit{A}) }  & \textit{S} & 1.000 & 1.000 & 1.000 & - \\
      & $\lambda^{[l]}$ & 2.90 & 2.89 & 2.41 & 0.83\\
      \midrule 
      \multirow{2}{*}{VGG-Biased (\textit{B})} & \textit{S} & 1.000 & 1.000 & 1.000 & -\\
      & $\lambda^{[l]}$ & 2.24 & 3.25 & 2.61 & 0.69\\
      \midrule 
      \multirow{2}{*}{VGG-Biased (\textit{C})} & \textit{S} & 1.000 & 1.000 & 1.000 & -\\
      & $\lambda^{[l]}$ & 2.36 & 2.52 & 2.86 & 0.82\\
      \midrule 
      VGG-Unbiased & \textit{S} & 1.000 & 1.000 & 1.000 & -\\
       & $\lambda^{[l]}$ & 2.49 & 2.67 & 2.51 & 0.93\\
      \midrule 
      \midrule 
      \multirow{2}{*}{ResNet-Biased (\textit{A}) }  & \textit{S} & 1.000 & 1.000 & 1.000 & -\\
      & $\lambda^{[l]}$ & 2.82 & 2.65 & 2.53 & 0.90\\
      \midrule 
      \multirow{2}{*}{ResNet-Biased (\textit{B})} & \textit{S} & 0.999 & 1.000 & 1.000 & -\\
      & $\lambda^{[l]}$ & 2.11 & 2.47 & 2.35 & 0.85\\
      \midrule 
      \multirow{2}{*}{ResNet-Biased (\textit{C})} & \textit{S} & 1.00 & 1.000 & 1.000 & -\\
     & $\lambda^{[l]}$ & 2.11 & 2.32 & 2.35 & 0.90\\
     \midrule 
      ResNet-Unbiased & \textit{S} & 0.999 & 0.999 & 0.999 & -\\
       & $\lambda^{[l]}$ & 2.33 & 2.32 & 2.34 & 0.99\\
      \midrule 
      \bottomrule 
    \end{tabular}
    \end{center}
\label{tabla_activation}
\end{table}

\section{Conclusions}\label{conclusions}

In this work we presented a preliminary analysis of how biased 
data affect the learning processes of deep neural network 
architectures in terms of activation level. We showed how ethnic attributes affect the learning
process of gender classifiers. We evaluated these differences in
terms of filter activation, besides performance, and the results
showed how the biases are encoded heavily in the last layers of 
the models. This activation reveals behaviors usually hidden 
during the learning process. We also evaluated different training
strategies that suggest to what extent biases can be reduced if the whole 
network is trained using a heterogeneous dataset. 

We finally propose a novel method, InsideBias, to detect bias through layer activations.
InsideBias has two major advantages with respect to detection based on
performance differences across demographic groups: 1) it does not require 
many samples (we showed that biased behaviors can be detected with only 15 images),
and 2) InsideBias can give an indication of the bias in the model 
using only good samples correctly recognized (even with the highest confidence).

\section{Acknowledgments}
\label{ack}
This work has been supported by projects BIBECA (RTI2018-101248-B-I00 MINECO/FEDER), TRESPASS (MSCA-ITN-2019-860813), PRIMA (MSCA-ITN-2019-860315), and Accenture. I. Serna is supported by a research fellowship from the Spanish CAM.




\bibliographystyle{IEEEtran}
\bibliography{IEEEabrv,AAAI.bib}

\end{document}